\documentclass[letterpaper]{article} 
\usepackage{aaai2026}  
\usepackage{times}  
\usepackage{helvet}  
\usepackage{courier}  
\usepackage[hyphens]{url}  
\usepackage{graphicx} 
\usepackage{natbib}  
\usepackage{caption} 
\usepackage{algorithm}
\usepackage{graphicx}
\usepackage{amsmath}
\usepackage{amssymb}
\usepackage{colortbl}
\usepackage{graphicx}
\usepackage{xcolor}
\usepackage{multirow}
\usepackage{booktabs} 
\usepackage{amssymb}
\usepackage{pifont}

\usepackage{algpseudocode}
\usepackage{newfloat}
\usepackage{listings}
\DeclareCaptionStyle{ruled}{labelfont=normalfont,labelsep=colon,strut=off} 
\floatstyle{ruled}
\newfloat{listing}{tb}{lst}{}
\floatname{listing}{Listing}
\title{GOAL: Geometrically Optimal Alignment for Continual \\ Generalized Category Discovery}
\author{
    Jizhou Han\textsuperscript{\rm 1},
    Chenhao Ding\textsuperscript{\rm 2},
    SongLin Dong\textsuperscript{\rm 3},
    Yuhang He\textsuperscript{\rm 1}\thanks{Corresponding author.},\\
    Shaokun Wang\textsuperscript{\rm 4},
    Qiang Wang\textsuperscript{\rm 1},
    Yihong Gong\textsuperscript{\rm 1}
}

\affiliations{
    \textsuperscript{\rm 1} National Key Laboratory of Human-Machine Hybrid Augmented Intelligence,\\National Engineering Research Center for Visual Information and Applications,\\Institute of Artificial Intelligence and Robotics, Xi'an Jiaotong University \\
    \textsuperscript{\rm 2} School of Software Engineering, Xi'an Jiaotong University, \\
    \textsuperscript{\rm 3} Shenzhen University of Advanced Technology, \\
    \textsuperscript{\rm 4} Harbin Institute of Technology, Shenzhen

     jizhou-han@stu.xjtu.edu.cn, heyuhang@xjtu.edu.cn, ygong@mail.xjtu.edu.cn
}

\usepackage{bibentry}

\begin{document}

\maketitle

\begin{abstract}
Continual Generalized Category Discovery (C-GCD) requires identifying novel classes from unlabeled data while retaining knowledge of known classes over time. Existing methods typically update classifier weights dynamically, resulting in forgetting and inconsistent feature alignment. We propose GOAL, a unified framework that introduces a fixed Equiangular Tight Frame (ETF) classifier to impose a consistent geometric structure throughout learning. GOAL conducts supervised alignment for labeled samples and confidence-guided alignment for novel samples, enabling stable integration of new classes without disrupting old ones. Experiments on four benchmarks show that GOAL outperforms the prior method \textit{Happy}, reducing forgetting by {16.1\%} and boosting novel class discovery by {3.2\%}, establishing a strong solution for long-horizon continual discovery.
\end{abstract}

\section{Introduction}
\label{sec:intro}

\begin{figure*}[h]
  \centering
   \includegraphics[width=0.99\linewidth]{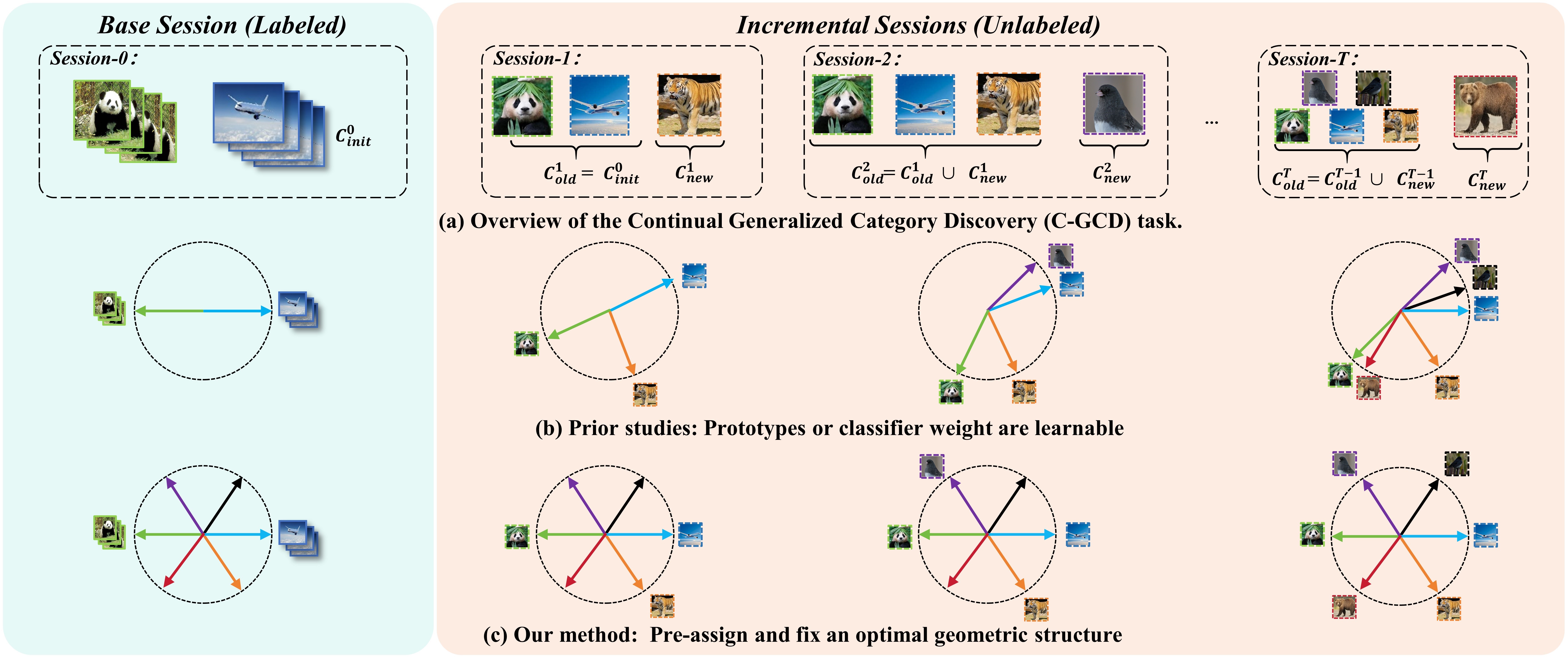}
   \caption{Conceptual illustration of the Continual Generalized Category Discovery (C-GCD) task and classifier design strategies. (a) Overview of the C-GCD task: a model is trained on labeled base classes and incrementally discovers novel categories from unlabeled data in subsequent sessions. (b) Prior studies typically use learnable prototypes or classifier weights. (c) Our method adopts a fixed optimal geometric structure, ensuring better alignment and generalization across continual sessions.}
   \label{fig:intro}
\end{figure*}

Continual Generalized Category Discovery (C-GCD) aims to incrementally identify novel categories from unlabeled data while retaining recognition of previously learned classes. Unlike conventional semi-supervised learning, which assumes all categories are predefined and labeled, C-GCD operates in a more realistic open-world setting. In this scenario, only a small set of known classes is labeled in the initial stage, and subsequent sessions provide unlabeled data that include a mixture of known and novel categories. The model must generalize from limited supervision to new concepts while avoiding forgetting of learned knowledge.

Recent efforts in GCD and C-GCD have led to notable advancements. Early approaches~\cite{GCD2022} used contrastive learning to separate known and unknown classes. Building on this idea, DCCL~\cite{DCCL2023} and PromptCAL~\cite{PromptCAL2023} introduced dynamic contrastive objectives and prompt-based affinities to refine decision boundaries. Other methods~\cite{PIM2023,CMS2024} enhanced representation quality via clustering and mean-shift refinement. In the continual setting, MetaGCD~\cite{metagcd2023} leveraged meta-initialization to improve cross-session transfer, and Happy~\cite{happy2024} mitigated bias through entropy-based regularization and prototype sampling.

However, most existing C-GCD frameworks rely on dynamically optimized classifier heads or class prototypes, which present two major challenges. \textbf{First, catastrophic forgetting}: The continuous update of class representations across sessions causes newly learned knowledge to overwrite prior information, particularly when novel classes are trained in a purely unsupervised manner. \textbf{Second, category confusion}: Without an explicit geometric constraint on the feature space, similar or nearby classes—especially emerging ones—tend to overlap, resulting in ambiguous decision boundaries. As illustrated in Fig.~\ref{fig:intro}b, existing methods often struggle to maintain global class separability.
These observations lead to a natural question:
\textbf{\textit{Can we predefine an optimal and unified geometric structure for all classes, enabling consistent feature alignment throughout continual discovery while alleviating forgetting and class confusion?}}

We draw inspiration from the Neural Collapse phenomenon~\cite{Papyan2020_NeuralCollapse}, which reveals that well-trained classifiers tend to organize features in an Equiangular Tight Frame (ETF), where class means are maximally separated and intra-class variation collapses. This structure provides an optimal configuration for classification, both theoretically and empirically.

Inspired by the Neural Collapse phenomenon, we propose a novel framework named Geometrically Optimal Alignment for Continual Generalized Category Discovery (GOAL), which establishes a globally consistent classification structure by utilizing a fixed ETF, as depicted in Fig. 1c. In contrast to existing approaches that iteratively adjust classifier prototypes across sessions, GOAL predefines an optimal geometric configuration and enforces alignment of both labeled and unlabeled features to this fixed structure throughout the entire continual learning process. The framework is composed of a frozen ETF-based classifier, a learnable feature encoder, and a confidence-guided alignment module. During the base session, labeled samples are explicitly guided to align with their designated ETF directions, thereby forming an initial structured representation space. In the subsequent incremental sessions, unlabeled data containing both previously seen and novel categories are processed by evaluating the prediction confidence of each sample. A fixed proportion of samples with the highest confidence, identified through minimum entropy, are selected and aligned with unassigned ETF vectors. This mechanism facilitates the stable integration of novel categories while preserving the global geometric consistency of the feature space.
By maintaining a consistent geometric objective across sessions, GOAL enhances class separability, stabilizes feature alignment, and effectively mitigates forgetting in the absence of explicit supervision for novel categories.
Our main contributions are summarized as follows:
\begin{itemize}
\item We propose GOAL, a Neural Collapse-inspired framework for continual generalized category discovery, which leverages a fixed ETF classifier to provide a unified and consistent optimization target across all sessions.
\item We design a confidence-guided alignment strategy that incrementally assigns high-confidence novel samples to unoccupied ETF directions, preserving feature structure without requiring ETF classifier prototypes updates.
\item GOAL achieves strong performance on four C-GCD benchmarks, reducing average forgetting by \textbf{16.10\%} and improving novel category discovery accuracy by \textbf{3.19\%} compared to the strongest prior method~\cite{happy2024}.
\end{itemize}

\section{Related Work}
\label{sec:RW}

\subsection{Generalized Category Discovery}

Generalized Category Discovery (GCD) aims to classify known categories while discovering novel ones from partially labeled data. Early works such as VanillaGCD~\cite{GCD2022} introduced contrastive learning to distinguish between known and unknown classes. DCCL~\cite{DCCL2023} and PromptCAL~\cite{PromptCAL2023} further refined decision boundaries via dynamic similarity modeling and prompt-based affinity propagation.  
Parametric solutions like SimGCD~\cite{SimGCD2023} and GPC~\cite{GPC2023} employed Gaussian mixture modeling or soft classifiers to capture latent structures. Others, including PIM~\cite{PIM2023}, CMS~\cite{CMS2024}, and AGCD~\cite{AGCD2024}, emphasized representation refinement through clustering and  mean-shift mechanisms. NCGCD \cite{han2025unleashing} explores the application of ETF in GCD, but it cannot perform incremental learning.
Despite their effectiveness, these methods update prototypes during training, introducing optimization inconsistency that hampers the discovery of novel categories.

\subsection{Continual Generalized Category Discovery }
Continual Generalized Category Discovery (C-GCD) extends the GCD setting to continual learning, where the model receives data in sequential sessions containing a mixture of known and unknown classes.  
GM~\cite{GM2022} proposes a grow-and-merge pipeline, clustering new classes and merging them with prior knowledge. MetaGCD~\cite{metagcd2023} leverages meta-learning to initialize models for better cross-stage generalization. Happy~\cite{happy2024} tackles prediction bias via entropy-based regularization and prototype resampling. 
However, current C-GCD methods still face challenges, including \textbf{catastrophic forgetting} and the lack of a \textbf{global geometric constraint} to preserve inter-class structure across sessions. These issues lead to inconsistent optimization and degraded performance over time.

\subsection{Neural Collapse and Geometric Structures}

Neural Collapse (NC)~\cite{Papyan2020_NeuralCollapse} characterizes the terminal phase of deep classifier training, where within-class features collapse to their means, and class means form a Simplex Equiangular Tight Frame (ETF). This configuration achieves maximum inter-class margin and minimal intra-class variance.  
NC has been theoretically shown to emerge under both cross-entropy~\cite{Weinan2020_Tetrahedral,Fang2021_LayerPeeled,Graf2021_DissectingContrastive,Wang2025DualCP} and MSE~\cite{Mixon2020_UnconstrainedNC,Zhou2022_ForwardFSCIL,Han2022_NeuralCollapseMSE} losses. Moreover, NC-inspired formulations have been applied to domains such as imbalanced classification~\cite{zhong2023understanding}, federated learning~\cite{Huang2023_NCFederatedLearning}, and few-shot class-incremental learning~\cite{Yang2023_DynamicSupportFSCIL}.

In this work, we are the first to introduce an NC-guided geometric constraint into the C-GCD setting. By assigning a fixed ETF classifier and aligning features through both supervised and confident unsupervised signals, we ensure a \textbf{globally consistent optimization target}, which enhances stability, improves novel category separability, and mitigates forgetting across continual sessions.

\begin{figure*}[t]
  \centering
  \includegraphics[width=0.99\linewidth]{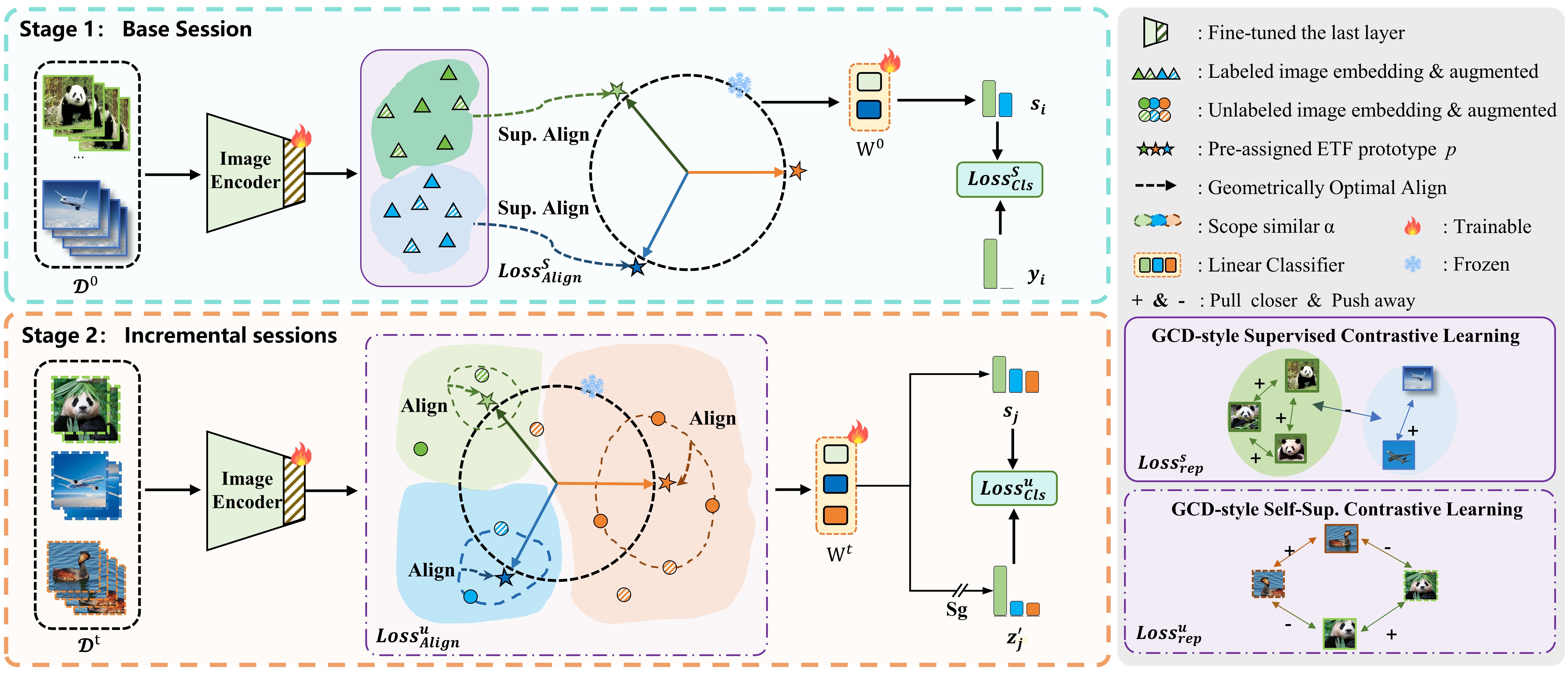}
    \caption{Overview of the GOAL framework. In the base session, labeled features are aligned to fixed ETF prototypes via supervised training. In incremental sessions, high-confidence unlabeled samples are aligned to unallocated prototypes with unsupervised learning. This maintains geometric consistency for continual discovery and retention.}
  \label{fig:main}
\end{figure*}

\section{Preliminaries}
\label{sec:Preliminaries}

\subsection{Problem Setting of C-GCD}

Continual Generalized Category Discovery (C-GCD) extends the conventional GCD task to a continual learning scenario. Formally, the learning process is divided into an initial supervised session (Stage-0) and a sequence of incremental discovery sessions (Stage-$t$, $1 \leq t \leq T$).

In Stage-0, the model is trained on a small labeled dataset $\mathcal{D}_{\text{train}}^0 = \{(x_i^l, y_i)\}_{i=1}^{N^0}$, where $y_i \in \mathcal{C}_{\text{init}}^0$ and $\mathcal{C}_{\text{old}}^0 = \mathcal{C}_{\text{init}}^0$ denotes the set of initially known categories.

For each incremental stage $t \in \{1, 2, ..., T\}$, the model receives an unlabeled dataset $\mathcal{D}_{\text{train}}^t = \{x_i^u\}_{i=1}^{N^t}$, which contains instances sampled from a mixture of known and unknown categories. The full category set at stage $t$ is defined as $\mathcal{C}^t = \mathcal{C}_{\text{old}}^t \cup \mathcal{C}_{\text{new}}^t$, where $\mathcal{C}_{\text{old}}^t$ denotes previously discovered categories and $\mathcal{C}_{\text{new}}^t$ represents novel categories appearing at this stage. The number of novel classes $K_{\text{new}}^t = |\mathcal{C}_{\text{new}}^t|$ is assumed to be either given or estimated.
After each stage, the model is evaluated on a disjoint test set $\mathcal{D}_{\text{test}}^t = \{(x_i, y_i)\}_{i=1}^{N_{\text{test}}^t}$ that covers all known and newly discovered categories.

\subsection{Definition of Optimal Geometric Structure}

In the final stages of training well-regularized neural networks, especially on balanced classification tasks, a remarkable structural phenomenon known as \textit{Neural Collapse} emerges~\cite{Papyan2020_NeuralCollapse}. This behavior reveals that both feature embeddings and classifier weights organize into a highly symmetric configuration, where samples from the same class collapse to a shared center, and class centers are arranged in an optimal geometric layout. This structure is mathematically characterized by a \textit{Simplex Equiangular Tight Frame (ETF)}.

\noindent\textbf{Simplex ETF Construction.}
A set of $K$ prototype vectors $P = [p_1, p_2, \dots, p_K] \in \mathbb{R}^{d \times K}$ forms a simplex ETF if they satisfy the following condition:
\begin{equation}
    P = \sqrt{\frac{K}{K - 1}} U \left(I_K - \frac{1}{K} \mathbf{1}_K \mathbf{1}_K^\top \right),
    \label{eq:ETF}
\end{equation}
where $U \in \mathbb{R}^{d \times K}$ is an orthonormal matrix such that $U^\top U = I_K$, and $\mathbf{1}_K$ is a column vector of ones. Each vector $p_k$ is unit-normalized, i.e., $\|p_k\| = 1$, and the pairwise inner product between any two vectors satisfies:
\begin{equation}
    p_k^\top p_j = 
    \begin{cases}
    1 & \text{if } k = j \\
    -\frac{1}{K - 1} & \text{if } k \neq j
    \end{cases}
\end{equation}
which ensures equal angular separation between all classes.

\noindent\textbf{Neural Collapse Properties.}
The Neural Collapse phenomenon includes four key behaviors:

\noindent\textit{(NC1) Within-class Collapse}: All features from the same class converge to their class mean. The within-class scatter tends to vanish, i.e., $\Sigma_W^{(k)} \rightarrow 0$, where $\Sigma_W^{(k)} = \text{Avg} \left\{ (\mu_{k, i} - \mu_k)(\mu_{k, i} - \mu_k)^T \right\}$ and $\mu_{k,i}$ is the feature of the $i$-th instance in class $k$.

\noindent\textit{(NC2) Simplex Configuration}: After centering each class mean by subtracting the global mean $\mu_G = \text{Avg}_{i,k}(\mu_{k, i})$, the normalized class means $\hat{\mu}_k = \frac{\mu_k - \mu_G}{\|\mu_k - \mu_G\|}$ tend to occupy the vertices of a simplex ETF.

\noindent\textit{(NC3) Feature-Classifier Alignment}: The centered class means align with the corresponding normalized classifier weights, i.e., $\hat{\mu}_k = \frac{w_k}{\|w_k\|}$, where $w_k$ is the weight vector for class $k$ in the final linear classifier.

\noindent\textit{(NC4) Prediction Simplification}: In this regime, classification can be interpreted as a nearest-mean decision rule: $\hat{y} = \arg\min_k \|\mu - \mu_k\| = \arg\max_k \langle \mu, w_k \rangle$, where $\mu$ is the test feature representation.

\begin{table*}[t]
\small
\setlength{\tabcolsep}{2.65pt}
\renewcommand{\arraystretch}{1}
\centering
\label{tab:full_gcd_results}
\begin{tabular}{ll c|ccc|ccc|ccc|ccc|ccc|>{\columncolor{gray!10}}c>{\columncolor{gray!10}}c>{\columncolor{gray!10}}c}
\toprule
\multirow{2}{*}{{Datasets}} 
& \multirow{2}{*}{{Methods}} 
& \multicolumn{1}{c}{S-0} 
& \multicolumn{3}{c}{{Stage-1}} 
& \multicolumn{3}{c}{{Stage-2}} 
& \multicolumn{3}{c}{{Stage-3}} 
& \multicolumn{3}{c}{{Stage-4}} 
& \multicolumn{3}{c}{{Stage-5}} 
& \multicolumn{3}{c}{\cellcolor{gray!10}{1-5 Stages Avg}} \\
 \cmidrule(lr){3-3} 
\cmidrule(lr){4-6} \cmidrule(lr){7-9} \cmidrule(lr){10-12} 
\cmidrule(lr){13-15} \cmidrule(lr){16-18} \cmidrule(lr){19-21}
& & \multicolumn{1}{c}{All} 
& {All} & {Old} & \multicolumn{1}{c}{New} 
& {All} & {Old} & \multicolumn{1}{c}{New} 
& {All} & {Old} & \multicolumn{1}{c}{New} 
& {All} & {Old} & \multicolumn{1}{c}{New} 
& {All} & {Old} & \multicolumn{1}{c}{New} 
& {All} & {Old} & {New} \\

\midrule
\multirow{9}{*}{\rotatebox{45}{C100}}
& VanillaGCD  \textdagger & 90.8 & 72.3 & 78.5 & 41.4 & 67.0 & 72.5 & 34.3 & 58.0 & 62.3 & 28.1 & 56.6 & 59.5 & 33.0 & 51.4 & 53.7 & 30.3 & 61.1 & 65.3 & 33.4 \\
& SimGCD \textdagger & 90.4 & 73.4 & 86.4 & 8.0 & 62.6 & 72.4 & 3.3 & 54.2 & 61.6 & 2.1 & 47.6 & 53.4 & 1.6 & 43.5 & 47.9 & 4.6 & 56.2 & 64.3 & 3.9 \\
& SimGCD+  \textdagger& 90.4 & 75.9 & \textbf{87.0} & 20.4 & 67.1 & 75.3 & 17.5 & 58.4 & 64.3 & 17.3 & 54.3 & 58.7 & 19.1 & 50.5 & 53.9 & 19.8 & 61.3 & 67.9 & 18.8 \\
& FRoST  \textdagger& 90.4 & 76.9 & 79.6 & \underline{63.3} & 65.3 & 68.9 & 43.9 & 58.0 & 61.1 & 36.5 & 49.3 & 50.9 & 36.2 & 48.0 & 48.2 & 46.8 & 59.5 & 61.7 & 45.3 \\
& GM \textdagger & 90.4 & 76.6 & 79.8 & 60.5 & 71.1 & 74.5 & \underline{50.6} & 63.5 & 68.2 & 31.0 & 59.7 & 62.5 & 37.6 & 54.1 & 54.7 & \underline{48.4} & 65.0 & 68.0 & 45.6 \\
& MetaGCD \textdagger & {90.8} & 76.1 & 83.6 & 38.7 & 69.4 & 72.8 & 48.9 & 62.0 & 65.8 & 35.3 & 58.2 & 61.2 & 34.3 & 55.8 & 58.5 & 31.6 & 64.3 & 68.4 & 37.8 \\
& Happy  \textdagger& 90.4 & \underline{80.4} & 85.3 & 56.1 & \underline{74.1} & \underline{78.3} & 49.3 & \underline{68.2} & \underline{70.9} & \underline{49.8} & \underline{62.3} & \underline{63.8} & \textbf{50.3} & \underline{60.0} & \underline{61.0} & \textbf{51.3} & \underline{69.0} & \underline{71.8} & \underline{51.4} \\
\rowcolor{gray!15} \cellcolor{gray!0} & \textbf{GOAL (Ours)} & {91.0} & \textbf{82.8} & \underline{86.4} & \textbf{65.1} & \textbf{77.8} & \textbf{84.0} & \textbf{62.4} & \textbf{71.8} & \textbf{82.8} & \textbf{53.5} & \textbf{66.6} & \textbf{83.8} & \underline{45.1} & \textbf{61.5} & \textbf{79.2} & 43.8 & \textbf{72.1} & \textbf{83.2} & \textbf{54.0} \\

\midrule
\multirow{9}{*}{\rotatebox{45}{Tiny}} 
& VanillaGCD  \textdagger & 84.2 & 55.9 & 58.9 & 41.0 & 55.0 & 58.6 & 33.2 & 52.8 & 55.7 & 32.4 & 48.8 & 51.5 & 27.6 & 45.9 & 48.1 & 26.9 & 51.7 & 54.6 & 32.2 \\
& SimGCD  \textdagger & 85.9 & 67.0 & 79.9 & 2.0 & 57.8 & 67.0 & 2.8 & 52.7 & 59.8 & 2.8 & 45.0 & 50.3 & 2.8 & 41.6 & 45.8 & 3.8 & 52.8 & 60.6 & 2.8 \\
& SimGCD+ \textdagger & 85.9 & 70.4 & {81.8} & 13.3 & 62.5 & 70.8 & 12.8 & 54.6 & 60.5 & 13.2 & 48.0 & 52.5 & 11.9 & 43.0 & 46.5 & 12.7 & 55.7 & 62.4 & 12.8 \\
& FRoST \textdagger  & 85.9 & 75.2 & 78.6 & \underline{58.1} & 65.6 & 67.8 & \textbf{52.5} & 51.3 & 54.3 & 30.4 & 48.2 & 52.1 & 16.9 & 40.2 & 42.7 & 16.9 & 56.1 & 59.1 & 35.0 \\
& GM  \textdagger & 85.9 & {76.4} & {82.4} & 46.5 & \underline{68.9} & \underline{73.8} & 39.2 & 58.7 & 63.4 & 25.4 & 52.9 & 57.2 & 18.1 & 46.9 & 50.6 & 13.4 & 60.8 & 65.5 & 28.5 \\
& MetaGCD \textdagger  & 84.2 & 60.9 & 64.9 & 40.8 & 57.2 & 61.0 & 34.2 & 54.4 & 57.2 & 34.6 & 50.8 & 53.6 & 28.8 & 48.1 & 50.2 & 30.0 & 54.3 & 57.4 & 33.7 \\
& Happy \textdagger  & {85.9} & \textbf{78.9} & \textbf{82.4} & \textbf{61.1} & 71.3 & 76.2 & 42.3 & \underline{64.7} & \underline{68.7} & \underline{36.5} & \underline{58.5} & \underline{60.6} & \underline{41.3} & \underline{54.6} & \underline{56.7} & \underline{35.7} & \underline{65.6} & \underline{68.9} & \underline{43.4} \\

\rowcolor{gray!15} \cellcolor{gray!0} & \textbf{GOAL (Ours)} & {86.1} &  \underline{77.1} &  \underline{81.9} & 52.8 & \textbf{72.1} & \textbf{80.3} & \underline{51.6} & \textbf{67.3} & \textbf{77.0} & \textbf{51.1} & \textbf{61.5} & \textbf{76.4} & \textbf{{42.8}} & \textbf{57.2} & \textbf{74.5} & \textbf{39.9} & \textbf{67.1} & \textbf{78.1} & \textbf{47.9} \\

\midrule
\multirow{9}{*}{\rotatebox{45}{CUB}} 
& VanillaGCD \textdagger & 89.2 & 64.5 & 67.1 & 51.9 & 58.2 & 60.7 & 42.9 & 54.1 & 56.4 & 37.9 & 50.0 & 51.3 & 39.3 & 46.8 & 46.6 & \underline{49.1} & 54.7 & 56.4 & 44.2 \\
& SimGCD \textdagger & {90.3} & 73.8 & 84.5 & 22.0 & 63.4 & 72.4 & 8.6 & 55.6 & 62.0 & 11.1 & 49.3 & 54.6 & 7.9 & 44.7 & 48.7 & 9.3 & 57.4 & 64.4 & 11.8 \\
& SimGCD+ \textdagger & 90.3 & 75.6 & \textbf{85.6} & 26.0 & 65.3 & 73.9 & 13.7 & 57.4 & 63.3 & 16.3 & 51.1 & 55.7 & 14.3 & 45.8 & 49.3 & 14.3 & 59.1 & 65.6 & 16.9 \\
& FRoST \textdagger & 90.3 & 77.0 & 84.0 & 43.5 & 50.8 & 53.5 & 34.3 & 46.4 & 49.3 & 26.1 & 39.4 & 41.5 & 23.1 & 34.6 & 35.1 & 29.5 & 49.6 & 52.7 & 31.3 \\
& GM \textdagger & 90.3 & 76.2 & 80.2 & 56.5 & 67.9 & 73.4 & 34.6 & 61.1 & 66.5 & 23.0 & 55.9 & 57.5 & 43.4 & 52.0 & 54.4 & 30.1 & 62.6 & 66.4 & 37.5 \\
& MetaGCD \textdagger & 89.2 & 67.1 & 70.2 & 51.9 & 60.8 & 61.0 & 50.9 & 57.5 & 59.3 & 37.8 & 51.9 & 52.2 & 49.4 & 49.6 & 50.0 & 46.4 & 57.4 & 58.8 & 47.3 \\
& Happy \textdagger & 90.3 & \textbf{81.4} & \underline{85.1} & \underline{63.7} & \textbf{74.3} & \underline{76.0} & \underline{63.6} & \underline{67.1} & \underline{71.1} & 39.1 & \underline{62.3} & \underline{63.8} & \underline{49.7} & \underline{59.4} & \underline{60.5} & \textbf{49.5} & \underline{68.9} & \underline{71.3} & \underline{53.1} \\
 \rowcolor{gray!15} \cellcolor{gray!0} & \textbf{GOAL (Ours)} & {90.4} & \underline{81.1} & 83.1 & \textbf{71.3} & \underline{73.1} & \textbf{76.4} & \textbf{65.0} & \textbf{69.0} & \textbf{76.1} & \textbf{57.3} & \textbf{64.2} & \textbf{74.6} & \textbf{51.4} & \textbf{61.9} & \textbf{76.0} & 47.9 & \textbf{69.9} & \textbf{77.2} & \textbf{58.6} \\

\midrule
\multirow{9}{*}{\rotatebox{45}{IN100}} 
& VanillaGCD  \textdagger & 96.0 & 70.1 & 72.9 & 56.2 & 69.4 & 73.5 & 44.8 & 68.5 & 70.6 & 53.6 & 65.6 & 67.8 & 47.2 & 64.5 & 67.4 & 38.4 & 67.6 & 70.5 & 48.0 \\
& SimGCD  \textdagger& 96.2 & 79.7 & 91.7 & 19.6 & 70.2 & 78.8 & 18.6 & 61.9 & 67.4 & 23.2 & 56.7 & 60.9 & 22.6 & 52.9 & 56.4 & 21.4 & 64.3 & 71.1 & 21.1 \\
& SimGCD+  \textdagger& 96.2 & 83.1 & 95.2 & 22.6 & 74.6 & 83.5 & 21.2 & 67.6 & 73.6 & 25.8 & 62.1 & 66.8 & 24.2 & 57.6 & 61.5 & 23.0 & 69.0 & 76.1 & 23.4 \\
& FRoST \textdagger & 96.2 & 87.5 & 93.0 & 60.2 & 79.6 & 83.4 & 57.2 & 76.8 & 77.0 & \underline{75.2} & 66.2 & 68.7 & 46.4 & 63.8 & 66.4 & 40.6 & 74.8 & 77.7 & 55.9 \\
& GM \textdagger & 96.2 & 89.5 & 95.0 & 62.0 & 82.3 & 86.9 & 54.8 & 78.0 & 79.2 & 69.6 & 72.8 & 74.7 & 58.0 & 71.1 & 71.8 & 65.0 & 78.7 & 81.5 & 61.9 \\
& MetaGCD \textdagger & 96.0 & 75.3 & 78.2 & 60.6 & 73.8 & 75.9 & 54.9 & 69.4 & 72.2 & 49.4 & 67.2 & 70.1 & 44.2 & 66.7 & 69.3 & 43.0 & 70.5 & 73.2 & 50.4 \\
& Happy  \textdagger& 96.2 & \underline{91.2} & \underline{95.4} & \underline{70.4} & \underline{87.8} & \underline{90.8} & \underline{69.8} & \underline{85.2} & \underline{86.4} & \textbf{77.0} & \textbf{81.9} & \underline{83.0} & \textbf{73.4} & \underline{78.6} & \underline{79.1} & \textbf{73.8} & \underline{85.0} & \underline{86.9} & \underline{72.9} \\
 \rowcolor{gray!15} \cellcolor{gray!0}  & \textbf{GOAL (Ours)} & {96.2} & \textbf{93.9} & \textbf{95.9} & \textbf{83.8} & \textbf{88.8} & \textbf{94.9} & \textbf{73.7} & \textbf{86.4} & \textbf{94.4} & 72.9 & \underline{81.2} & \textbf{92.4} & \underline{68.4} & \textbf{79.3} & \textbf{93.0} & \underline{66.6} & \textbf{85.9} & \textbf{94.1} & \textbf{73.1} \\
\bottomrule
\end{tabular}
\caption{Performance of C-GCD on four benchmarks. Here \textdagger denotes results reported by \textit{Happy}.}
\label{tab:full_gcd_results}

\end{table*}

\section{Method}
\label{sec:method}

\subsection{Overview}

As illustrated in Fig.~\ref{fig:main}, our GOAL framework addresses Continual Generalized Category Discovery (C-GCD) by unifying supervised and unsupervised alignment under a fixed geometric structure. GOAL comprises three key components: a frozen ETF prototype set \( P \), a feature encoder \( f(\cdot) \), and a session-specific classifier \( W \). These modules are optimized jointly across sessions without storing exemplars.

In the base session, the model is trained with labeled data to establish a structured feature space. Feature embeddings are aligned to fixed ETF directions via dot-product loss, while a supervised contrastive loss further enhances intra-class compactness. A parametric classifier \( W^0 \) is also trained to support known-class recognition.

In each incremental session (\( t \geq 1 \)), the model receives unlabeled data containing both seen and novel categories. High-confidence samples are selected via entropy ranking and aligned to unused ETF directions. A clustering-guided strategy is used to initialize new classifier weights, and a pseudo-label-based classification loss guides training. Additionally, unsupervised contrastive learning further improves feature discrimination.
By maintaining alignment with a global ETF structure throughout all sessions, GOAL enables progressive discovery of novel classes while preserving knowledge of previous ones—achieving effective C-GCD without any memory buffer or prototype updates.

\subsection{Base Session}
\label{sec:base_session}
The base session uses labeled data to build a structured feature space by aligning features with predefined ETF prototypes and learning discriminative representations. This establishes a unified geometric foundation for recognizing known classes and discovering novel ones in later sessions.

\subsubsection{Predefined ETF Prototypes}

To promote a well-structured feature space and ensure maximum separation between categories, we construct a fixed set of classifier prototypes based on an ETF. Given a total of \(K\) categories, the corresponding ETF prototype matrix is denoted as \( \mathbf{P} = \{p_1, p_2, \dots, p_K\} \in \mathbb{R}^{d \times K} \), as defined in Eq.~\ref{eq:ETF}.
Each prototype \(p_i\) is normalized to unit \(\ell_2\)-norm and satisfies a fixed pairwise cosine similarity:
\begin{equation}
    p_i^\top p_j = \frac{K}{K - 1} q_i^\top q_j - \frac{1}{K - 1}, \quad \forall i, j \in [1, K],
\end{equation}
where \(q_i\) denotes the \(i\)-th column of an orthonormal matrix.

This construction provides a stable and optimal alignment structure, serving as a consistent geometric target for both known and novel categories.

\subsubsection{Supervised Alignment to ETF Prototypes}

For each labeled training sample $(x_i^l, y_i^l) \in \mathcal{D}_{\text{train}}^0$, we extract a feature embedding $e_i = f(x_i^l)$ using the image encoder $f(\cdot)$. The feature is then normalized to unit length: $\hat{e}_i = e_i / \|e_i\|_2$, and we align it with the pre-assigned ETF prototype $p_{y_i^l}$ corresponding to the ground-truth label $y_i^l$.
The alignment loss is:
\begin{equation}
    \mathcal{L}_{\text{Align}}^{s} = - \frac{1}{N^0} \sum_{i=1}^{N^0} \langle \hat{e}_i, p_{y_i^l} \rangle,
\end{equation}
where $\langle  \rangle$ denotes inner product and $N^0$ is the number of labeled samples in the base session.

\subsubsection{Representation Learning and Parametric Classification}

\textit{Contrastive Representation Learning.}
We adopt a GCD-style representation learning \cite{GCD2022}. 
For each image \( x_i^l \in \mathcal{D}_{\text{train}}^0 \), we generate two augmented views \( x_i \) and \( x_i' \), and extract their normalized embeddings \( e_i \) and \( e_i' \) using the encoder \( f(\cdot) \). The unsupervised contrastive loss is:
\begin{equation}
  \mathcal{L}_{\text{rep}}^{u} = -\frac{1}{|B|} \sum_{i \in B} \log \frac{\exp(e_i \cdot e_i' / \tau)}{\sum_{j \in B, j \neq i} \exp(e_i \cdot e_j / \tau)},
\end{equation}
\( \tau \) is a temperature hyperparameter and \( B \) is the batch size.

For labeled samples, we further encourage semantic consistency through a supervised contrastive loss:
\begin{equation}
\mathcal{L}_{\text{rep}}^{s} = -\frac{1}{|B_l|} \sum_{i \in B_l} \frac{1}{|H(i)|} \sum_{h \in H(i)} 
\log \frac{\exp(e_i \cdot e_h / \tau)}{\sum_{j \neq i} \exp(e_i \cdot e_j / \tau)},
\end{equation}
where \( H(i) \) is the set of positive samples sharing the same label with \( i \). The total contrastive loss is:
\begin{equation}
  \mathcal{L}_{\text{rep}}^{Base} = (1 - \lambda_{\text{rep}}) \mathcal{L}_{\text{rep}}^{u} + \lambda_{\text{rep}} \mathcal{L}_{\text{rep}}^{s}.
\end{equation}

\noindent \textit{Parametric Classification.}
We also employ a trainable parametric classifier \( W^0 \in \mathbb{R}^{d \times K} \) initialized for the known categories. For labeled samples, the classification loss is defined as standard cross-entropy:
\begin{equation}
  \mathcal{L}_{\text{cls}} = -\frac{1}{|B_l|} \sum_{i \in B_l} \log \frac{\exp(w_{y_i}^\top e_i)}{\sum_{j=1}^K \exp(w_j^\top e_i)},
\end{equation}
where \( w_j \) is the weight vector for class \( j \), and \( y_i \) is the true label for sample \( x_i \).
The total training objective for the base session is:
\begin{equation}
  \mathcal{L}_{Base} = \mathcal{L}_{{Align}}^{s} + \mathcal{L}_{\text{rep}}^{Base} + \mathcal{L}_{\text{cls}}.
\end{equation}

\begin{table*}[t]
\small
\setlength{\tabcolsep}{7.5pt}
\renewcommand{\arraystretch}{1}
\centering
\label{tab:forget_discovery}
\begin{tabular}{l cc| cc| cc |cc |>{\columncolor{gray!10}}c>{\columncolor{gray!10}}c}
\toprule
\multirow{2}{*}{{Methods}} 
& \multicolumn{2}{c}{{CIFAR100}} 
& \multicolumn{2}{c}{{CUB-200}} 
& \multicolumn{2}{c}{{TinyImageNet}} 
& \multicolumn{2}{c}{{ImageNet-100}} 
& \multicolumn{2}{c}{{\cellcolor{gray!10}All Datasets Avg}} \\
\cmidrule(lr){2-3} \cmidrule(lr){4-5} \cmidrule(lr){6-7} \cmidrule(lr){8-9} \cmidrule(lr){10-11}
& \( \mathcal{M}_f \downarrow \) & \multicolumn{1}{c}{\( \mathcal{M}_d \uparrow \) }
& \( \mathcal{M}_f \downarrow \) & \multicolumn{1}{c}{\( \mathcal{M}_d \uparrow \) }
& \( \mathcal{M}_f \downarrow \) & \multicolumn{1}{c}{\( \mathcal{M}_d \uparrow \) }
& \( \mathcal{M}_f \downarrow \) & \multicolumn{1}{c}{\( \mathcal{M}_d \uparrow \) }
& \( \mathcal{M}_f \downarrow \) & \( \mathcal{M}_d \uparrow \) \\
\midrule
FRoST (ECCV 2022) & 42.19 & 45.34 & 55.14 & 31.30 & 43.13 & 34.96 & 29.80 & 55.92 & 42.57 & 41.88 \\
GM (NeurIPS 2022) & 35.62 & {45.62} & 35.86 & 37.51 & 35.24 & 28.52 & 24.44 & 61.88 & 32.79 & 43.38 \\
MetaGCD (CVPR 2023) & 32.35 & 37.76 & 39.24 & {47.27} & 34.04 & {33.68} & 26.65 & 50.42 & 33.07 & {42.28} \\
Happy (NeurIPS 2024) & \underline{29.40} & \underline{51.36} & \underline{29.77} & \underline{53.13} & \underline{29.20} & \underline{43.38} & \underline{17.09} & \underline{72.88} & \underline{26.37} & \underline{55.19} \\
\rowcolor{gray!15} \textbf{GOAL (Ours)} & \textbf{11.82} & \textbf{53.97} & \textbf{14.36} & \textbf{58.57} & \textbf{11.66} & \textbf{47.88} & \textbf{3.24} & \textbf{73.08} & \textbf{10.27} & \textbf{58.38} \\
\rowcolor{gray!25}  \textbf{\textit{Improv. over Happy}} & \textit{\textbf{+17.58}} & \textit{\textbf{+2.61}} & \textit{\textbf{+15.41}} & \textit{\textbf{+5.44}} & \textit{\textbf{+17.54}} & \textit{\textbf{+4.50}} & \textit{\textbf{+13.85}} & \textit{\textbf{+0.20}} & \textit{\textbf{+16.10}} & \textit{\textbf{+3.19}} \\
\bottomrule
\end{tabular}
\caption{Forgetting and discovery performance on four datasets. We report forgetting rate (\( \mathcal{M}_f \)) and final discovery rate (\( \mathcal{M}_d \)).}
\label{tab:forget_discovery}
\end{table*}

\begin{table*}[t]
\centering
\small
\setlength{\tabcolsep}{2.5pt}
\renewcommand{\arraystretch}{1.0}
\label{tab:c100_stagewise}
\begin{tabular}{llccccccccccc|ccc|c}
\toprule
\multirow{2}{*}{Datasets} & \multirow{2}{*}{Methods} 
& S-0 & S-1 & S-2 & S-3 & S-4 & S-5 & S-6 & S-7 & S-8 & S-9
&  \multicolumn{1}{c}{S-10} & \multicolumn{3}{c}{1-10-stages Avg} & \multirow{2}{*}{$M_f$ $\downarrow$} \\
\cmidrule(lr){3-3} \cmidrule(lr){4-4} \cmidrule(lr){5-5} \cmidrule(lr){6-6} \cmidrule(lr){7-7} \cmidrule(lr){8-8} \cmidrule(lr){9-9} \cmidrule(lr){10-10} \cmidrule(lr){11-11} \cmidrule(lr){12-12} \cmidrule(lr){13-13} \cmidrule(lr){14-16}
& & All & All & All & All & All & All & All & All & All & All
& \multicolumn{1}{c}{All} & All & Old & \multicolumn{1}{c}{New} & \\
\midrule
\multirow{3}{*}{{C100}} 
& MetaGCD    & {90.82} & 81.07 & 76.55 & 74.26 & 67.64 & 64.45 & 61.58 & 59.13 & 60.13 & 56.91 & 56.51 & 65.82 & 68.49 & 31.82 & 32.81 \\
& Happy      & 90.36 & \underline{85.62} & \underline{81.88} & \underline{79.82} & \underline{74.01} & \underline{71.81} & \underline{68.46} & \underline{64.05} & \underline{62.14} & \underline{61.38} & \underline{57.81} & \underline{70.70} & \underline{71.89} & \underline{52.34} & \underline{31.00} \\
\rowcolor{gray!15}
 \cellcolor{gray!0} & \textbf{GOAL (Ours)} & 90.60 & \textbf{87.22} & \textbf{84.50} & \textbf{83.20} & \textbf{77.66} & \textbf{75.55} & \textbf{73.72} & \textbf{70.07} & \textbf{68.04} & \textbf{66.48} & \textbf{63.06} & \textbf{74.95} & \textbf{82.63} & \textbf{62.62} & \textbf{11.86} \\
\rowcolor{gray!25}
\multicolumn{2}{c}{\textit{\textbf{Improv. over Happy}}}
& \textbf{\textit{+0.24}} & \textbf{\textit{+1.60}} & \textbf{\textit{+2.62}} & \textbf{\textit{+3.38}} & \textbf{\textit{+3.65}} & \textbf{\textit{+3.74}} & \textbf{\textit{+5.26}} & \textbf{\textit{+6.02}} & \textbf{\textit{+5.90}} & \textbf{\textit{+5.10}} & \textbf{\textit{+5.25}} & \textbf{\textit{+4.25}} & \textbf{\textit{+10.74}} & \textbf{\textit{+10.28}} & \textbf{\textit{+19.14}} \\

\midrule
\multirow{3}{*}{{Tiny}} 
& MetaGCD    & 84.20 & 68.87 & 65.48 & 62.92 & 60.81 & 58.21 & 56.16 & 54.56 & 52.48 & 50.57 & 48.92 & 57.90 & 59.91 & 30.76 & \underline{34.35} \\
& Happy      & \underline{85.86} & \underline{80.75} & \underline{76.92} & \underline{73.34} & \underline{69.77} & \underline{66.33} & \underline{62.75} & \underline{57.56} & \underline{54.73} & \underline{53.02} & \underline{50.69} & \underline{64.59} & \underline{66.24} & \underline{43.04} & 34.91\\
\rowcolor{gray!15} \cellcolor{gray!0}
& \textbf{GOAL (Ours)} & \textbf{86.14} & \textbf{80.96} & \textbf{79.23} & \textbf{76.43} & \textbf{72.40} & \textbf{68.85} & \textbf{64.95} & \textbf{62.28} & \textbf{58.63} & \textbf{56.77} & \textbf{53.03} & \textbf{67.35} & \textbf{79.15} & \textbf{47.23} & \textbf{11.20} \\
\rowcolor{gray!25}
\multicolumn{2}{c}{\textit{\textbf{Improv. over Happy}}}
& \textbf{\textit{+0.28}} & \textbf{\textit{+0.21}} & \textbf{\textit{+2.31}} & \textbf{\textit{+3.09}} & \textbf{\textit{+2.63}} & \textbf{\textit{+2.52}} & \textbf{\textit{+2.20}} & \textbf{\textit{+4.72}} & \textbf{\textit{+3.90}} & \textbf{\textit{+3.75}} & \textbf{\textit{+2.34}} & \textbf{\textit{+2.77}} & \textbf{\textit{+12.91}} & \textbf{\textit{+4.19}} & \textbf{\textit{+23.71}} \\
\bottomrule
\end{tabular}
\caption{C-GCD performance with 10 incremental stages on CIFAR100 and TinyImageNet.}
\label{tab:c100_stagewise}

\end{table*}

\subsection{Incremental Sessions}
\label{sec:incremental_session}

After the base session, the model proceeds through a series of incremental sessions \( t \in \{1, 2, \dots, T\} \), where it receives unlabeled data \( \mathcal{D}_{\text{train}}^t = \{x_j^u\}_{j=1}^{N^t} \) consisting of both known and novel categories. The challenge lies in discovering new classes while maintaining performance on old ones, without access to labeled exemplars.

\subsubsection{Parametric Classification}

To handle new classes, we expand the classifier using clustering-guided initialization~\cite{happy2024}. Specifically, we apply KMeans clustering on the unlabeled dataset \( \mathcal{D}_{\text{train}}^t \) to obtain a set of \(\ell_2\)-normalized cluster centers \( \{c_i\} \). For each cluster center \( c_i \), we compute its maximum cosine similarity with old classifier weights \( W_{t-1} \). We then select the top \( C_{\text{new}} \) centers with the lowest similarity scores to serve as new class weights \(W_t^{new}= c_{t_j} \).
The classifier weights at stage \( t \) are obtained by concatenating the old and new weights:
\begin{equation}
    W_t = Concat [W_{t-1}; W_t^{new}] \in \mathbb{R}^{d \times (C_{\text{old}} + C_{\text{new}})}.
\end{equation}

For training, each unlabeled sample $x_j^u$ is passed through the image encoder and the classifier to get a prediction $\tilde{s}_j$, and its augmented view $x_j^{u'}$ provides a sharper target distribution $z_j$. The classification loss is:
\begin{equation}
  \mathcal{L}_{\text{cls}}^{u} = \frac{1}{|\mathcal{D}_{\text{train}}^t|} \sum_{x_j \in \mathcal{D}_{\text{train}}^t} \mathcal{H}(z_j, \tilde{s}_j) - \epsilon \mathcal{H}(\bar{s}),
\end{equation}
$\mathcal{H}$ denotes cross-entropy, \( \epsilon \) is a regularization parameter and $\bar{s}$ is the mean batch prediction.

Finally, we rank all samples by prediction entropy and select the top $\alpha\%$ with lowest entropy to form the confident subset $\mathcal{D}_{\text{conf}}^t$ for ETF alignment in the next step.

\subsubsection{Unsupervised ETF Alignment}
Following the confidence guided strategy, we cluster the selected high-confidence samples \( \mathcal{D}_{\text{conf}}^t \) and match the resulting centroids to unused ETF prototypes. Each sample \( x_j \in \mathcal{D}_{\text{conf}}^t \) is assigned to a prototype \( p_{\phi(j)} \), and the alignment loss is defined as:
\begin{equation}
  \mathcal{L}_{\text{Align}}^{u} = - \frac{1}{|\mathcal{D}_{\text{conf}}^t|} \sum_{x_j \in \mathcal{D}_{\text{conf}}^t} \langle \hat{e}_j, p_{\phi(j)} \rangle,
\end{equation}
where \( \hat{e}_j \) is the normalized embedding and \( \phi(j) \) is the index of the assigned ETF prototype. This mechanism anchors novel features to the global structure.

\subsubsection{Incremental Objective}
To improve representation quality under unlabeled conditions, we retain unsupervised contrastive learning for each sample, denoted as \( \mathcal{L}_{\text{rep}}^{Inc} \). The overall objective is:
\begin{equation}
  \mathcal{L}_{\text{Inc}} = \lambda_A \mathcal{L}_{\text{Align}}^{u} + \mathcal{L}_{\text{rep}}^{Inc} + \mathcal{L}_{\text{cls}}^{u},
\end{equation}
where \( \lambda_A \) balances alignment strength.

\newcommand{\cmark}{\checkmark} 
\newcommand{\xmark}{\ding{55}}  
\section{Experiment}

\subsection{Experimental Setup}

\noindent \textbf{Datasets.} 
We evaluate GOAL on four C-GCD benchmarks: CIFAR100~\cite{krizhevsky2009learning}, TinyImageNet~\cite{deng2009imagenet}, ImageNet-100~\cite{tian2020contrastive}, and CUB~\cite{wah2011caltech}. Following~\cite{GCD2022}, each dataset is split into a labeled base session (50\% classes) and \( T = 5 \) incremental sessions with unlabeled novel classes and previously seen classes.

\noindent \textbf{Evaluation Protocol.}
At each stage \( t \), the model is trained on unlabeled training data \( \mathcal{D}_{\text{train}}^t \) and evaluated on a disjoint labeled test set \( \mathcal{D}_{\text{test}}^t \). We adopt an inductive evaluation, where ground-truth labels are only used for computing accuracy. Predictions are matched to true classes using the Hungarian algorithm over all classes seen so far. We report accuracy for three subsets: “All” (overall), “Old”, and “New”.

\noindent \textbf{Implementation Details.}
We use a pre-trained DINO ViT-B/16~\cite{caron2021emerging, dosovitskiy2021image} as the backbone, fine-tuning only the last block and a 3-layer MLP projection head (hidden size 2048, output dim 768). Stage-0 is trained for 100 epochs with labeled data; each incremental session runs for 30 epochs using a batch size of 128 and learning rate 0.01. We set \( \lambda_A = 0.7 \) and \( \lambda_{\text{rep}} = 0.35 \). All experiments are conducted on NVIDIA RTX 3090 GPU.

\begin{table*}[t!]
\centering
\small
\setlength{\tabcolsep}{4.5pt}
\begin{tabular}{cc ccc ccc ccc ccc ccc}
\toprule
\multirow{2}{*}{\shortstack{Sup ETF\\Alignment}} & \multirow{2}{*}{\shortstack{Unsup ETF\\Alignment}} & 
\multicolumn{3}{c}{CUB} & 
\multicolumn{3}{c}{CIFAR100} & 
\multicolumn{3}{c}{TinyImageNet} & 
\multicolumn{3}{c}{ImageNet-100} & 
\multicolumn{3}{c}{Avg} \\
\cmidrule(r){3-5} \cmidrule(r){6-8} \cmidrule(r){9-11} \cmidrule(r){12-14} \cmidrule(r){15-17}
 &  & All & Old & New & All & Old & New & All & Old & New & All & Old & New & All & Old & New \\
\midrule
\xmark & \xmark & 64.8 & 73.1 & 49.9 & 69.6 & 79.9 & 47.5 & 63.3 & 72.5 & 44.2 & 80.1 & 87.9 & 68.0 & 69.5 & 78.4 & 52.4 \\
\cmark & \xmark & 64.7 & \textbf{77.5} & 51.1 & 70.4 & \textbf{83.7} & 46.2 & 63.4 & \underline{73.7} & 44.7 & 82.3 & \textbf{94.3} & 68.7 & 70.2 & \underline{82.3} & 52.7 \\
\xmark & \cmark & \underline{67.9} & 73.8 & \underline{57.3} & \underline{71.2} & 81.2 & \underline{52.4} & \underline{65.6} & 72.8 & \underline{46.1} & \underline{84.7} & 89.5 & \underline{72.1} & \underline{72.4} & 79.3 & \underline{57.0} \\
\rowcolor{gray!20} \cmark & \cmark & \textbf{69.9} & \underline{77.2} & \textbf{58.6} & \textbf{72.1} & \underline{83.2} & \textbf{54.0} & \textbf{67.1} & \textbf{78.1} & \textbf{47.9} & \textbf{85.9} & \underline{94.1} & \textbf{73.1} & \textbf{73.8} & \textbf{83.2} & \textbf{58.4} \\
\rowcolor{gray!40} \multicolumn{2}{c}{\textit{Improv. over baseline}} & 
\textbf{+5.1} & \textbf{+4.1} & \textbf{+8.7} & 
\textbf{+2.5} & \textbf{+3.3} & \textbf{+6.5} & 
\textbf{+3.8} & \textbf{+5.6} & \textbf{+3.7} & 
\textbf{+5.8} & \textbf{+6.2} & \textbf{+5.1} & 
\textbf{+4.3} & \textbf{+4.8} & \textbf{+6.0} \\
\bottomrule
\end{tabular}
\caption{Ablation study on the effect of supervised and unsupervised ETF alignment.}
\label{tab:etf_ablation_ours}
\end{table*}

\begin{figure*}[t!]
  \centering
    \includegraphics[width=0.97\linewidth]{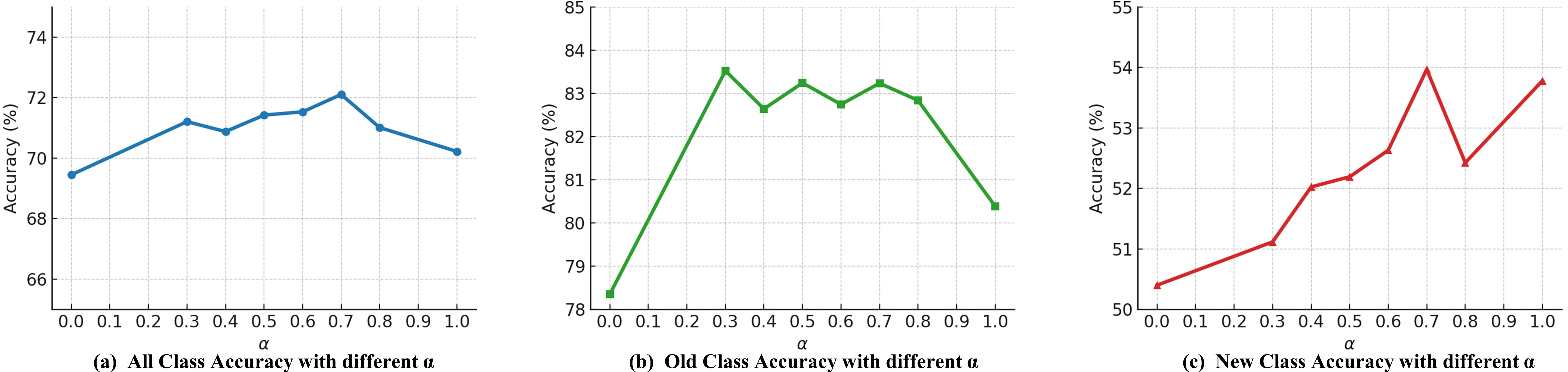}
  \caption{Accuracy trend across different values of \( \alpha \) on CIFAR100.}
  \label{fig:alpha_ablation}
\end{figure*}

\subsection{Comparison with State of the Art Methods}

We compare GOAL with representative approaches in Generalized Category Discovery (GCD) and its continual variant (C-GCD). These include VanillaGCD~\cite{GCD2022}, SimGCD~\cite{SimGCD2023}, SimGCD+ (with LwF), FRoST~\cite{frost2022}, GM~\cite{GM2022}, MetaGCD~\cite{metagcd2023}, and Happy~\cite{happy2024}. To ensure fairness, we follow MetaGCD's protocol for GM and use sampled features instead of exemplars. Full results are reported in Table~\ref{tab:full_gcd_results}, Table~\ref{tab:forget_discovery}, and Table~\ref{tab:c100_stagewise}.

As summarized in Table~\ref{tab:full_gcd_results}, GOAL establishes a new state of the art in C-GCD across all benchmarks. Our method achieves consistent and significant gains over the best prior approach, Happy. On CIFAR100, GOAL improves overall accuracy by +3.1\%, with clear benefits for both old +11.4\% and new +2.6\% classes. On TinyImageNet, GOAL narrows the old-new performance gap and yields a +4.5\% boost for novel categories. For CUB, GOAL significantly enhances new class discovery +5.5\% while maintaining strong retention. These results confirm the effectiveness of our confidence-driven alignment and fixed geometric structure in supporting balanced, continual category discovery.

\subsubsection{Forgetting and Discovery Performance.}

To evaluate the model’s continual learning ability, we adopt two metrics: the forgetting rate (\( \mathcal{M}_f \)) and the final discovery rate (\( \mathcal{M}_d \)) in \cite{GM2022}. The forgetting rate measures how much performance on previously learned classes degrades over time, while the discovery rate reflects the model’s accuracy in identifying novel classes introduced in later sessions.

As shown in Table~\ref{tab:forget_discovery}, GOAL consistently outperforms prior methods, achieving both the lowest forgetting and highest discovery scores across all benchmarks. Specifically, compared to \textit{Happy}, GOAL reduces \( \mathcal{M}_f \) by more than \textbf{17\%} on CIFAR100 and TinyImageNet, while improving \( \mathcal{M}_d \) by +2.6\% and +4.5\%, respectively. On CUB and IN100, GOAL also achieves substantial gains in both metrics. Averaged across all datasets, GOAL achieves a \textbf{16.1\%} reduction in forgetting and a \textbf{3.2\%} improvement in discovery, demonstrating its strong capability to retain old knowledge while continuously discovering new categories.

\noindent \textbf{C-GCD with More Continual Stages.}
We evaluate GOAL under a 10 stage setting on CIFAR100 and TinyImageNet. As shown in Table~\ref{tab:c100_stagewise}, GOAL improves new class accuracy over \textit{Happy} by \textbf{+10.28\%} and \textbf{+4.19\%}, respectively, while reducing forgetting by \textbf{19.14\%} and \textbf{23.71\%}. These results highlight GOAL’s strength in discovering novel categories and mitigating forgetting, making it especially suitable for long-horizon continual discovery.

\subsection{Ablation Study}

\noindent \textbf{Effect of ETF Alignment.}
To assess the contributions of \textit{Supervised} and \textit{Unsupervised ETF Alignment}, we evaluate four configurations: (1) Baseline without alignment, (2) supervised only, (3) unsupervised only, and (4) both enabled. Results in Table~\ref{tab:etf_ablation_ours} show consistent gains from both modules.

\noindent\textit{Unsupervised ETF Alignment.}
This component notably boosts novel class discovery, improving new class accuracy by +4.6\% on average. The improvement is especially prominent on fine-grained datasets, where modeling inter-class similarity is challenging. It guides confident samples toward unused ETF prototypes, enhancing discriminability.

\noindent\textit{Supervised ETF Alignment.}
This module improves retention, enhancing old class accuracy by +3.9\%. By aligning labeled features to fixed ETF directions, it anchors known categories and reduces forgetting. While its effect on novel class accuracy is smaller, the joint use of both alignment strategies yields the best overall performance.

\noindent \textbf{Influence of the $\alpha$ Coefficient.}  
We examine the effect of the ETF scope coefficient $\alpha$ on CIFAR100. As shown in Fig.~\ref{fig:alpha_ablation}, increasing $\alpha$ consistently improves novel class accuracy, reaching 53.97\% at $\alpha$ = 0.7. All-class accuracy also peaks at 72.10\% when $\alpha$ = 0.7, while old class accuracy remains stable across the range. These results indicate that a proper choice of $ \alpha$ enhances novel discovery without affecting old class retention.

\section{Conclusion and Future Work}
In this paper, we propose GOAL, a unified framework for Continual Generalized Category Discovery. By aligning features to fixed ETF prototypes and guiding novel class discovery through confidence-aware selection, GOAL achieves consistent improvements in both accuracy and stability across incremental sessions. Extensive experiments on four benchmarks demonstrate its superior performance in new class discovery and forgetting mitigation, especially under longer task horizons.
In future work, we will explore adaptive ETF expansion for dynamically estimating the number of novel classes and extend the framework to multimodal settings with textual or semantic cues.

\bibliography{aaai2026}

\end{document}